\documentclass{article}

\usepackage{arxiv}

\usepackage[utf8]{inputenc} 
\usepackage[T1]{fontenc}    
\usepackage{hyperref}       
\usepackage{url}            
\usepackage{booktabs}       
\usepackage{amsfonts}       
\usepackage{nicefrac}       
\usepackage{microtype}      
\usepackage{lipsum}		
\usepackage{graphicx}
\usepackage{natbib}
\usepackage{doi}
\usepackage{comment}
\usepackage{tikz}
\usepackage{pgfplots}
\usepackage{amsmath}
\usepackage{subfigure}

\title{Anchor Prediction: A Topic Modeling Approach}

\author{
	Jean Dupuy \\
	Université de Lyon, Lyon 2, ERIC EA3083\\
	MeetSYS\\
	Lyon, France \\
	\texttt{jean.dupuy@meetsys.com} \\
	\And
	Adrien Guille \\
	Université de Lyon, Lyon 2, ERIC EA3083\\
	Lyon, France \\
	\texttt{adrien.guille@univ-lyon2.fr} \\
	\And
	Julien Jacques \\
	Université de Lyon, Lyon 2, ERIC EA3083\\
	Lyon, France \\
	\texttt{julien.jacques@univ-lyon2.fr} \\
	}

\renewcommand{\shorttitle}{Anchor Prediction: A Topic Modeling Approach}

\begin{document}
\maketitle

\begin{abstract}
	
Networks of documents connected by hyperlinks, such as Wikipedia, are ubiquitous. Hyperlinks are inserted by the authors to enrich the text and facilitate the navigation through the network. However, authors tend to insert only a fraction of the relevant hyperlinks, mainly because this is a time consuming task. In this paper we address an annotation, which we refer to as \textit{anchor prediction}. Even though it is conceptually close to link prediction or entity linking, it is a different task that require developing a specific method to solve it. Given a source document and a target document, this task consists in automatically identifying anchors in the source document, \textit{i.e} words or terms that should carry a hyperlink pointing towards the target document. We propose a contextualized relational topic model, CRTM, that models directed links between documents as a function of the local context of the anchor in the source document and the whole content of the target document. The model can be used to predict anchors in a source document, given the target document, without relying on a dictionary of previously seen mention or title, nor any external knowledge graph. Authors can benefit from CRTM, by letting it automatically suggest hyperlinks, given a new document and the set of target document to connect to. It can also benefit to readers, by dynamically inserting hyperlinks between the documents they're reading. Experiments conducted on several Wikipedia corpora (in English, Italian and German) highlight the practical usefulness of anchor prediction and demonstrate the relevancy of our approach.
\end{abstract}


\section{Introduction}

    Wikipedia is an online and collaborative encyclopedia which is one the most visited website of the world. This attractiveness leads to the creation of nearly 600 new pages everyday in the English version of the encyclopedia\footnote{https://en.wikipedia.org/wiki/Wikipedia:Statistics}, and other languages tend to follow this trend. In Wikipedia, articles are linked together through a network of hyperlinks.
    Hyperlinks allow to enrich text and facilitate horizontal reading, by giving access to additional information during the reading. Formally, a hyperlink is a directed link, from a word or term -- which we refer to as \textit{anchor} -- in a source document towards a target document. A hyperlink in a Wikipedia page will look like this: \texttt{[[World Wide Web|WWW]]}. On the left of the pipe we find the title of the page to which the link points, and on the right the words that will carry the hyperlink.
    In Wikipedia hyperlinks are manually inserted by contributors, following a very detailed set of rules\footnote{https://en.wikipedia.org/wiki/Wikipedia:Manual\_of\_Style/Linking}.
    This task is community driven but could be time consuming for contributors, and some relevant anchors may remain unconnected. Thus, finding ways to improve the connectivity between pages by automatically suggesting or maybe inserting hyperlinks in the pages would facilitate the access to information and knowledge. 
    As mentioned in \cite{smith2020keeping} the use of machine learning in contributors' workflow needs to engaged the community, whose practices may differ regarding community of interest or language version. 
    Finally, recommending anchors from a document to another specified one allow contextual hyperlink generation, which can be used by the reader to track terms relative to a specific concept during its navigation though Wikipedia.

\paragraph{Problem definition} 
    Given a source document and a target document, we address the issue of automatically identifying potential anchors in the source document, given a target document. We refer to this task as \textit{Anchor Prediction}. It is different from link prediction, where the target document isn't given, and where the locations of the anchors don't matter. It also differs from entity-linking where targets aren't documents but entries of an ontology. Finally anchor prediction is close to an annotation task, but unlike the latter it aims to identify words carrying hyperlinks between a pair of specifics document, rather than looking for probables mentions of an entire set of entities. 

\paragraph{Use cases}
    Predicting anchors can be beneficial for both contributors and readers. 
    Having wrote a new document (i.e. the source), rather than directly inserting hyperlinks, a contributor could simply specify a set of target documents, and let the system automatically locate the anchors and thus insert or recommend the hyperlinks. This set of document could also be build with any recommendation method.
    During a reading session, contextual hyperlinks could be automatically inserted in the document being read (i.e. the source document), pointing towards previously read documents (i.e. targets), thus helping the users contextualizing their reading.

\paragraph{Related work}

    A lot of attention has been devoted to developing means to infer latent links between documents \citep{yang2015tadw, bojchevski2018g2g, liu2018content, brochier2019global}. Prior work on this issue views a link as a connection between two entire documents, following the more general definition of link prediction in simple networks \citep{survey_link_prediction}. In other words, most of existing methods -- including the most recent ones \citep{hao2020inductive, zhang2020help, brochier2020idne}, with the notable exception of~\cite{brochier2007anchor} -- ignore the locations of the links, i.e. anchors, in the documents and cannot trivially be adapted to solve the anchor prediction task. Yet, they provide grounds to develop new methods suited to anchor prediction. 
    
    Other settings like automatic annotation of documents have been also studied in the last decades. An annotation method aims to find in a text the most relevant mentions of documents that belong to a predefined corpus (Wikipedia for the large majority), and returns pairs of anchors and linked documents. 
    Those systems are build upon a wide range of machine learning methods, including statistical analysis of existing anchors \citep{ferragina2010tagme, piccinno2014wat}, topic modeling \citep{han2012entity} or deep learning \citep{ganea2017deep, wu2019zero}.
    However those methods are mostly designed to identify anchors pointing to another documents by relying on dictionaries of previously seen mentions and documents title, or knowledge graphs or onthologies. 

\paragraph{Proposal}
    In this paper, we propose a new method based on a contextualized relational topic model, which we refer to as \textit{CRTM}. We generalize the Relational Topic Model, RTM \citep{chang2009rtm}, itself based upon the well-known Latent Dirichlet Allocation (LDA) topic model \citep{blei2003lda}. To incorporate the network structure in LDA, RTM models undirected links between documents as a function of the topics assigned to all the words in each document. In contrast, \textit{CRTM} models directed links between documents as a function of the topics assigned to the words surrounding the anchor (\textit{i.e.} its context) in the source document and the topics assigned to all the words in the target document. We further improve the link function by (i) incorporating an attention mechanism to better estimate the importance of each topic for anchor prediction and (ii) incorporating additional parameters to learn new hidden representations more fitted to anchor prediction. CRTM is computationally light, language agnostic and is solely trained on texts and existing anchors, without any use of external knowledge bases nor documents titles. 

\paragraph{Results.} 
    We assess the ability of CRTM to correctly predict anchors on several corpora made of English, Italian and German Wikipedia articles. We compare its performance with those of other variants of RTM to highlight its relevancy, and a widely used annotation tool. We also conduct an ablation study, which reveals that both our improvements to the link function are crucial to the performance of CRTM. 
    To connect our work with prior work on link prediction, we provide additional results on this task, incorporating recent baselines. We show that, even though not initially designed to solve this task, CRTM still manages to beat some baselines, being beaten only by the most recent and specialized methods for link prediction. 
    To conclude this paper, we show how to use CRTM in practice to predict anchors. We provide examples that show that CRTM can help in automatically detecting words or terms that should carry hyperlinks. 

\section{Related Work}

In this section we discuss the different parts of the literature we lean on. Firstly, because CRTM is a topic model, we start by reviewing prior work on topic modeling. Next, we briefly survey work on link prediction, a close task to anchor prediction. Lastly, we discuss work that leverage anchor links in other settings.

\subsection{Topic Modeling}

We can classify topic models in two categories: LDA-based models and neural topic models.

\subsubsection{LDA-based Topic Models}These methods model documents as topic mixtures, each word in a document being assigned to one topic. LDA \citep{blei2003lda} is a generalization of pLSA~\citep{plsa}. 
It is a generative model to infer latent topics to describe a set of documents $D$. For each document $d \in D$, LDA infers a vector $\theta_d$ of proportions over a small number $K$ of latent topics. The topics are characterized by a distribution $\beta$ over a fixed vocabulary $V$. Each word occurrence $w_{d, i}$ in a document $d$ is independently assigned a topic, encoded in a one-hot manner as $z_{d,i} \in \{0;1\}^K$, according to the distribution over words for this topic, $\beta_{\bullet,z_{d,i}}$. 
The Hidden Topic Markov Model (HTMM)~\citep{gruber2007HTMM} models the topics of all the words in a same document as a Markov Chain, effectively lifting the independence assumption between neighboring word occurrences. More precisely HTMM expects that all words in a same sentence share the same topic, and that consecutive sentences are more likely to have the same topic. 
Biterm Topic Model (BTM)~\citep{yan2013biterm} proposes to address a shortcoming of LDA, dealing with short texts, by modeling the generation of word co-occurrence patterns. By taking advantage of co-occurrence at the corpus-level, BTM solves the sparsity problem at the document level, which makes it suitable for short texts.

\subsubsection{Neural Topic Models.} With the advent of deep generative models and variational auto-encoders \citep{kingma2019introduction}, topic modeling has been cast into the deep learning framework.
NTM-R~\citep{ding2018coherence} relies on neural variational inference and pre-trained word embeddings, and is trained with an objective towards topic coherence. W-LDA~\citep{nan2019topic} proposes to use Wasserstein auto-encoders~\citep{tolstikhin2018wasserstein} to address the problem of topic disentanglement described in~\cite{bengio2013representation}, and enforces the Dirichlet prior. 
GraphBTM~\citep{zhu2018graphbtm}, based on the same assumption than BTM, represents biterms as a graph and uses a Graph Neural Network (GCN)~\citep{kipf2017semi} to extract topic mixtures from them.  
\cite{zhou2020neural} proposes the Graph Topic Model (GTM), which extracts topic mixtures by processing a word-document graph with a GNN, to further improve topic coherence. Note that the word-document graph is distinct from an explicit document-document graph, as it solely computed from word occurrences in the documents.

It should be noted that neural topic models, such as NTM-R, W-LDA, GraphBTM or GTM, only compute topic mixtures and do not explicitly assign a topic to each word occurrence, as opposed to LDA-based models. 

\subsection{Link Prediction}
Link prediction is a related task to the one we propose. We classify existing approaches in two categories: approaches based on topic modeling, and approaches based on node embedding.

\subsubsection{Methods based on Topic Modeling}

The Relation Topic Model (RTM) \citep{chang2009rtm} is an extension of LDA that incorporates explicit links between documents, so that it is suitable for link prediction. Here, we quickly develop the internal functioning of this model, since CRTM generalizes it. 
Given a set $L$ of explicit links between the documents in a corpus $D$, RTM extends LDA with the addition of a link function that models the likelihood of the links between two documents in terms of their topic proportions. This function, $\psi(y_{d, d'} = 1)$, with $y_{d, d'} \in \{0;1\}$ a binary variable that indicates whether $d$ and $d'$ are linked, parametrized by $\eta, \nu \in \mathbb{R}^{K}$, is defined as:
\begin{equation}
    \psi_{(d,d')}(y=1)=\exp \left(\eta^{\top}\left({\overline{\mathbf{z}}}_{d} \circ \overline{\mathbf{z}}_{d^{\prime}}\right)+\nu\right),
    \label{eq:rtmexp}
\end{equation}{}
where $\circ$ denotes the Hadamard product and  $\overline{\mathbf{z}}_{d}=\frac{1}{N_{d}} \sum_{i} z_{d, i}$ and $\overline{\mathbf{z}}_{d'}=\frac{1}{N_{d}} \sum_{i} z_{d', i}$ represents the proportions of latent topics in document $d$ and $d'$ respectively. Overall, the generative process of RTM goes as follows:
\begin{itemize}
    \item For each document $d$ in $D$:
    \begin{itemize}
        \item draw topic proportion for $d$ : $\theta_d | \alpha \sim Dirichlet\left(\alpha\right)$
        \item for each word occurrence $w_{d,i}$ in $d$:
        \begin{itemize}
            \item draw topic assignment : $z_{d,i} | \theta_d \sim Multi\left(\theta_d\right)$
            \item draw word : $w_{d,i} | z_{d,i} \sim Multi\left(\beta_{z_{d,i}}\right)$
        \end{itemize}
        
    \end{itemize}
    \item For each pair of documents $(d,d')\in L$:
    \begin{itemize}
        \item draw link indicator : $y_{d,n} \sim \psi\left(\cdot | \eta,\nu, \mathbf{z}_{d}, \mathbf{z}_{d^{\prime}}\right)$
    \end{itemize}
\end{itemize}
$Multi(.)$ is the multinomial distribution, $\alpha$ the Dirichlet parameter, $\mathbf{z}_{d}$ a matrix whose rows are the $z_{d,n}$ vectors.

RTM has been modified or extended in different ways. gRTM \citep{chen2013generalized} proposes to capture all pairwise topic relationships between documents and relies on a different estimation procedure based on collapsed Gibbs sampling with data augmentation. \cite{zhang2013sparse} proposes a non-probabilistic formulation of RTM to control the sparsity of document's topics. \cite{yang2016discriminative} incorporates in RTM the weighted  stochastic  block  model \citep{aicher2013adapting} to identify blocks of strongly connected documents. \cite{yang2015birds} proposes a joint model that uses link structure to define clusters of documents.
RTM has also triggered works in the field of Neural Topic Modeling.
For instance, Relational Deep Learning (RDL)~\citep{wang2017relational} is a deep hierarchical Bayesian model designed for link prediction. 
\cite{bai2018neural} proposes Neural Relational Topic Model (NRTM) to mimic the RTM architecture with components from deep learning. For the LDA part NRTM uses a stacked variational auto-encoder and a multi-layer perceptron to model the link function. 


\subsubsection{Node embedding with textual information}
These methods learn document embeddings in a Euclidean space and then straightforwardly predict links according to the dot-product between these embeddings. Several neural architectures have been explored to learn the embeddings from the documents' content and the network structure, such as Graph2Gauss~\citep{bojchevski2018g2g}, a deep energy-based encoder embedding each nodes as a Gaussian distribution, STNE, a self-translating recurrent network \citep{liu2018content} or IDNE, an attention-based network \citep{brochier2020idne}. Methods based on matrix-factorization have also been investigated, like TADW \citep{yang2015tadw}, an extension of the matrix formulation of DeepWalk that leverages both the network structure and the content of the documents. Similarly, GVNR-t \citep{brochier2019global}, adapts the GloVe algorithm to embed networks of documents. Some methods, like RLE~\citep{gourru2020document}, aims at projecting linked documents in a pre-trained word embedding space.

It should be noted that all these approaches model links at the document-level, ignoring anchors (i.e. the position of the hyperlinks in the source document).

\subsection{Leveraging anchors}
\label{sec:anchorsRW}
In this last section, we discuss works that incorporate anchors to solve annotation tasks.

While some topic models take advantage of the anchor information \citep{sun2008htm, gruber2008LTMH, gruber2008LALTMH}, those methods are not suitable for a anchor prediction task. However many methods are designed to annotate documents by identifying portions of text related to another document.
The first method to address this task WIKIFY \citep{mihalcea2007wikify}. 
TAGME \citep{ferragina2010tagme} is an efficient method designed to annotate shorts texts with Wikipedia entries, combining three steps. For a given text, TAGME first looks for all mentions of entries in a dictionary containing titles of Wikipedia pages and anchor texts. Then an entity disambiguation step is performed using a voting scheme. The score of each mention-entity pair is computed as the sum of votes given by candidate entities of all other mentions in the text, taking into account the semantic association between two entities\citep{mei2008topic} and the probability of an entity being the link target of a given mention\citep{medelyan2008topic}. This step return on anchor per mention. Finally all the candidates below a certain threshold of coherence are pruned. 
WAT \citep{piccinno2014wat} is another method that extends TAGME by redefining the disambiguation step using graph-based methodologies \citep{hoffart2011robust}. 
\cite{gerlach2021multilingual} proposes a linking system adapted for Wikipedia which keep contributors in the loop.
\cite{sun2015modeling} is one of the first deep-learning based method designed to deal with this task, by leveraging the semantics of mention, context and entity as well as their compositionality in a unified way. 
BLINK \citep{wu2019zero} is another deep learning based method using a two stages approach for entity linking and annotation, based on fine-tuned BERT architectures.

\section{Contextualized Relational Topic Model}\label{CRTM}

In this section, we describe our proposal: the Contextualized Relational Topic Model, which we refer to as CRTM.  It generalizes RTM so that links are modeled as a function of the topics assigned to the words in the context of the anchor in the source document and the topics assigned to all the words in the target document. The context is a set of words that surround the anchor in the text. It can be rather narrow, \textit{e.g.} only the anchor word or the sentence in which it appears, or wider, up to the all the words in the document. In the latter case, CRTM is equivalent to RTM. We further refine the link function by (i) incorporating an attention mechanism and (ii) introducing additional parameters.

\subsection{The CRTM Model}

Let $d$ and $d'$ be two documents, and $c = \{w_i\}_{i=1}^C$ be the context of the link, \textit{i.e.} a set of $C \geq 1$ words, in the source document $d$. The link function of \textit{CRTM} is defined as:
\begin{equation}
        \psi_{d,d'}(y=1 ; c)=\exp \left(\eta^{\top}\left({Q\overline{\mathbf{z}}}_{d,c} \circ Q\overline{\mathbf{z}}_{d^{\prime}}\right)+\nu\right),
    \label{eq:CRTM}
\end{equation}
where $\overline{\mathbf{z}}_{d, c}$ is an attention-based weighted average of the per-word topic assignments in $c$ and $Q \in \mathbb{R}^{K \times K}$ is a set of additional parameters. We describe these two components in the following sub-sections.

\subsubsection{Attention-based Weighted Average}

Rather than calculating $\overline{\mathbf{z}}_{d, c}$ as a simple average, we rely on pre-trained word embeddings to calculate a weighted average of the topic assignments in $c$ using an attention mechanism. Assuming a word embedding $u_i \in \mathbb{R}^p$ for each word $w_i \in V$, we compute the attention scores $s \in \mathbb{R}^k$, by measuring the scaled dot-product \citep{vaswani2017scale-dot-product} between the embedding of the word carrying the link, $u_{link}$ and the embedding $u_{j}$ of each word $w_j$ in $c$:
\begin{equation}
    s_j = \frac{u_{link} \cdot u_j}{\sqrt{p}}.
\end{equation}
With the attention weights $a = \text{softmax}(s)$, we calculate the weighted average:
\begin{equation}
    \overline{\mathbf{z}}_{d,c} = \sum_{w_j \in c} a_j z_{d,j},
\end{equation}
where $z_{d,j}$ is the one-hot encoding of the topic assignment for the word $w_j$ in context $c$. This way, the topics assigned to words that are semantically closer to the word that carries the link are given more importance.

\subsubsection{Additional Parameters}

The form of the link function in RTM implies that linked documents should have similar topic proportions. We argue that this implicit assumption is too restricting as complementary documents that exhibit different topic proportions could be linked. Hence, we incorporate additional parameters $Q \in \mathbb{R}^{K \times K}$ so that CRTM learns new representations of the documents $d$ and $d'$ as linear transforms of $\overline{\mathbf{z}}_{d, c}$ and $\overline{\mathbf{z}}_{d'}$: \textit{i.e.} $Q\overline{\mathbf{z}}_{d, c}$ and $Q\overline{\mathbf{z}}_{d'}$. This way, RTM can compare $d$ and $d'$ beyond simple same-topic interactions. Note that the size of $Q$ is dictated by the estimation procedure, because it requires $\eta$ to be a $K$-dimensional vector, which in turns enforces $Q\overline{\mathbf{z}}_{d, c}$ and $Q\overline{\mathbf{z}}_{d'}$ to be $K$-dimensional vectors.

\subsection{Parameter Estimation}
We estimate the parameters adapting the procedure in RTM, by maximizing the likelihood using the variational Expectation-Maximization algorithm \citep{bishop2006PRML}. 
The addition of $Q$ in $\psi$ slightly modifies the expression of the ELBO and therefore the update rules of the variational parameters of $\theta_d$, and parameters for $\psi$: $\eta$ and $\nu$. The expected value of the link function becomes:
\begin{equation}
\mathbb{E}_{q}\left[\log p\left(y_{d, d'}=1 \mid \overline{\mathbf{z}}_{d}, \overline{\mathbf{z}}_{d'}, \boldsymbol{\eta} \right)\right]=\boldsymbol{\eta}^{T} \left({Q\overline{\mathbf{\phi}}}_{d,c} \circ Q\overline{\mathbf{\phi}}_{d^{\prime}}\right)+\nu,\nonumber
\end{equation}
where $\phi_{d,n}$ is the variational parameter of $z_{d,n}$ and $\overline{\mathbf{\phi}}_{d} = \frac{1}{N_{d}} \sum_{n} \phi_{d, n}$. 

Parameters $\gamma_d$ and $\phi_{d,i}$, respectively the variational parameters of $\theta_d$ and $z_{d,i}$, are updated at the E-step as follow:
\begin{equation}
    \gamma_d = \alpha + \sum_i \phi_{d,i},
\end{equation}
and
\begin{equation}
\small
    \phi_{d,i} \propto \exp\left( \log \beta_{.,w_{d,i}} +  \mathbf{\Psi}(\gamma_d) - \mathbf{1}\mathbf{\Psi}(\mathbf{1}^T\gamma_d)) + \sum_{d'\neq d}\eta \circ \frac{Q^2 \overline{\phi}_{d,i}}{N_d}\right),
\end{equation}
with $\mathbf{\Psi}(.)$ the Digamma function, $\mathbf{1}$ a K-dimensional vector of ones, and $N_d$ the number of words in document $d$. 

During the M-step we update the model's parameters $\beta$, $\eta$, $\nu$ and $Q$, according to the new values of $\gamma_d$ and $\phi_{d,i}$.
The update rule for $\beta$ didn't change from LDA:
\begin{equation}
    \beta_{k,w} \propto \sum_d \sum_n \mathbf{1}(w_{d,n} = w)\phi_{d,n}^k.
\end{equation}

The link function's parameters $\eta$ and $\nu$ are updated as follow:
\begin{equation}
    \eta \xleftarrow[]{}\log \left( \overline{\Pi} \right) - \log \left(\overline{\Pi} +\rho\overline{\pi}_{\alpha} \right) -\mathbf{1}\nu,
\end{equation}
 and 
 \begin{equation}
     \nu \xleftarrow{} \log\left( L - \mathbf{1}^T\overline{\Pi}\right) - \log\left(\rho(1-\mathbf{1}^T\overline{\pi}_{\alpha}) + L - \mathbf{1}^T \overline{\Pi} \right),
 \end{equation}
 with  $\overline{\Pi} = \sum_{(d,d')} Q\overline{\phi_d}\circ Q \overline{\phi_{d'}}$, $\overline{\pi}_{\alpha} = Q\frac{\alpha}{1^T\alpha} \circ Q \frac{\alpha}{1^T\alpha}$ and $L$ the total number of observed links. $\rho$ denote the number of negative examples needed, following the regularization procedure provided by~\cite{chang2010hierarchical}. Instead of using a $L2$ regularizer, we introduce negative observations, where $y=0$. The observations are associated with a similarity $\overline{\pi}_{\alpha}$, the expected Hadamard product of any two documents given the Dirichlet prior of the model.
 
Finally the coefficients of $Q$ are updated with
\begin{equation}
\small
 Q_{i,j} \leftarrow Q_{i,j} + l \times \sum_{(d, d' \in L)}\frac{\eta_i}{K} \left(\overline{\phi}_{d,j}\sum_{n=1}^K Q_{i,n}\overline{\phi}_{d',n} + \overline{\phi}_{d',j}\sum_{n=1}^K Q_{i,n}\overline{\phi}_{d,n}\right)\end{equation}
, where $l$ is a learning rate. Note that updates to $Q$ are strictly positive, which could cause numerical instability. While we could constrain $Q$ to prevent its norm from monotonically increasing and rely on the Lagrange multiplier to estimate it, we choose to address this issue differently and simply normalize each row of $Q$ with its $L2$-norm after each update. This tricks has been shown to work well in \cite{xing2015normalized} for instance.

\subsection{Time Complexity}

Estimating the parameters of the LDA component of CRTM has a time complexity of $\mathcal{O}(|D| \times K \times N)$, with $|D|$ the number of documents, $K$ the number of topics and $N$ the average number of words per document. Estimating the parameters of the relational component of CRTM has a complexity of $\mathcal{O}(L \times K^2)$, where $L$ is the number of observed links in the network, and $K^2$ is engendered by the addition of $Q$ in the link function. The proposed model thus have an overall $\mathcal{O}(|D| \times K \times N + L \times K^2)$ time complexity.

\subsection{Relationship to gRTM}

The link function of gRTM \citep{chen2013generalized} incorporates a matrix $Q' \in \mathbb{R}^{K \times K}$, as follow:
\begin{equation}
        \psi_{d, d'}(y=1)=\sigma \left({\overline{\mathbf{z}}}_{d} Q' \overline{\mathbf{z}}_{d^{\prime}}\right),
        \label{eq:gRTM}
\end{equation}
where $\sigma$ can be either the sigmoid function or the exponential function, as in RTM. This formulation arises from the fact that $\eta^{\top}\left({\overline{\mathbf{z}}}_{d} \circ \overline{\mathbf{z}}_{d^{\prime}}\right) = {\overline{\mathbf{z}}}_{d} \text{diag}(\eta) \overline{\mathbf{z}}_{d^{\prime}}$. Rather than enforcing $Q'$ to be a diagonal matrix to stick to RTM, gRTM considers $Q'$ as a full matrix that allows capturing inter-topics interactions. This is quite different from what \textit{CRTM} does, as the purpose of $Q$ isn't to replace $\eta$ but rather to project both $\overline{\mathbf{z}}_{d, c}$ and $\overline{\mathbf{z}}_{d'}$ to a new comparison space to improve link prediction.

\section{Experiments} \label{XP}

First, we quantitatively evaluate CRTM against RTM, TAGME and variants of RTM on the task of anchor prediction. Then we highlight the importance of contextualizing the link function, and show that both refinements we bring to the function are crucial to the performance of CRTM. Then we evaluate CRTM on a link prediction task against RTM and recent baselines. This experiment is less a way to evaluate if our model can reach state of the art results, since CRTM is designed for anchor prediction, than assess if our modifications leads to a weaker generalization. Lastly, we illustrate how to use our model for anchor prediction in practice.

\subsection{Datasets}

To demonstrate the relevancy of our proposal we construct six datasets using Wikipedia.
Each dataset is made of introductory sections of articles belonging to a specific category. Articles are sampled in a snow ball fashion following the links, starting from the main page of the category. Furthermore, since we have to construct our own datasets, we can easily evaluate our work on different languages, while commonly used datasets tend to be in only in English. We consider three languages: English, Italian and German. For each language we extract a dataset related to the main category about physics (\textit{Physics, Fisica} and \textit{Physik}), and society (\textit{Society, Società} and \textit{Gesellschaft}). We choose these categories to take into account different editorial policies, writing styles and vocabularies. Table~\ref{tab_corpus_summary} summarizes the general properties of these datasets.

\begin{table*}[]
    \caption{Dataset properties.}\label{tab_corpus_summary}
    \centering
    \begin{tabular}{lrrrrrr}
 \toprule
 \textbf{Category}    & \textit{Physics} &  \textit{Society} & \textit{Fisica} & \textit{Società} & \textit{Physik} & \textit{Gesellschaft}\\
 \midrule
 \textbf{Language}        & English &  English & Italian & Italian& German& German\\
 \textbf{Nb of pages}   & 6327 & 14486 & 2254 & 3106&3438&12262\\
 \textbf{Nb of links}   & 18821 & 30749 & 7145 & 7093 &10727&35710\\
 \textbf{Average number of words per doc}   &183 & 194 & 121 & 130&120&111\\
 \textbf{Average number of sentences per doc}& 9.77 & 10.05 & 5.79 & 5.94&8.69&8.12\\
 \textbf{Average degree}& 5.9 &  4.2 & 6.3 &4.5 &6.2&5.8\\
 \bottomrule
    \end{tabular}
\end{table*}

\subsection{Tasks and metrics}

We consider two tasks. The first is an anchor prediction task, and consists in locating the words carrying hyperlinks, hidden during training, given a link between a source and a target documents. Based on the scores given by $\psi_{d,d'}$, we:
\begin{itemize}
\item Calculate the precision at $n$, \textit{i.e.}  the fraction of hidden links for which the word carrying the hyperlink is ranked among the top $n$ words.
\item Construct the ROC curve, to visually assess the overall capacity of the models to give high ranks to the word carrying the hidden hyperlinks.
\end{itemize}

The second is a link prediction task. We hide a percent of edges
and compare the cosine similarity between hidden pairs and negative examples of unconnected documents. We report Area Under the ROC Curve. 

\subsection{Methodology}
\subsubsection{Anchor Prediction}
We hide one link per document, which represents between 21\% to 33\% of the links depending on the connectivity of the datasets. The set of hidden links is split into a validation set, used to control the convergence during the training phase, and a test set, used to calculate the precision and construct the ROC curves.
We compare CRTM, with the context matching the sentence in which the link occurs, against \textit{ad hoc} variants of RTM. gRTM was considered then discarded because of its prohibitive runtime on large corpora.
\begin{itemize}
    \item \textbf{CRTM}: We define the context as the sentence in which the hyperlink occurs. After gridsearching we set the learning rate $l$ to $0.01$ in the update rule for $Q$, which yield the best results. Word embeddings are trained independently on each dataset using Skip-Gram with negative sampling (window size of 10 and 20 negative samples per true sample) \citep{mikolov2013negativesampling}. By training one word-embedding set per corpora, we did not introduce external knowledge to our model and remain fair to the other baselines.
    
    \item \textbf{RTM}: We train the model following \cite{chang2009rtm}, with the original link function in Eq.~\ref{eq:rtmexp}. However this link function doesn't allow ranking the words because it is independent to the link's location. Hence, during the test phase, we change RTM's link function to $\psi_{(d,d')}(y=1 ; w_i)=\exp (\eta^{\top}(z_{d,i} \circ \overline{\mathbf{z}}_{d^{\prime}})+\nu)$, \textit{i.e.} we replace $\overline{\mathbf{z}}_{d^{\prime}}$ with $z_{d,i}$, the topic assignment for word $i$. This trick allows us to measure the probability of each word $w_i$ to carry the link. Note that we are limited to $z_{d,i}$, as considering larger context at test time would require modifications akin those proposed in CRTM.
    
    \item \textbf{TAGME}: In order to evaluate TAGME we use the annotation service provided by the D4Science's API\footnote{https://sobigdata.d4science.org/web/tagme/tagme-help}. 
    For each hidden hyperlink we report if TAGME identify the right linked document and the corresponding anchor. As TAGME only output one result per linked document because of its disambiguation process, only P@1 is reported.
    We set the confidence score threshold at $0.1$, discarding all results below this value. This value is the smaller threshold recommended by the documentation.

\end{itemize}

We also consider simpler variants of CRTM for the purpose of the ablation study:
\begin{itemize}
     \item \textbf{CRTM}$_1$ (no context): We consider singleton contexts that consist only contain the word carrying the link.
     
    \item \textbf{CRTM}$_U$ (uniform weighting): We match the contexts with sentences and remove the attention mechanism by directly calculating $\overline{\mathbf{z}}_{d, c}$ as a simple average. Because this link function is doesn't take the exact location of the word into account, we again use the trick that consists in restricting the context to a single word at test time.
    
    \item \textbf{CRTM}$_P$ (positional weighting): We match the contexts with sentences and replace the attention weights with weights calculated according to a Gaussian smoothing centered on the position of the word carrying the link. The weight for a word $w_i$ in the context is calculated as $g_i = e^{-\frac{1}{2}\left(\frac{\text{dist}(i)}{\sigma} \right)^2}$, where $\text{dist}(i)$ is the distance between $w_i$ and $w_\text{link}$, the word carrying the link. $\sigma$ is the standard deviation. We set it to $\sigma=3$ by gridsearch.
    
    \item \textbf{CRTM}$_I$ (no representation learning): We match the contexts with sentences and remove the additional parameters by setting $Q = I_K$, with $I_K$ the identity matrix.
\end{itemize}

All models are trained with 50 topics, which we've found to give the best performance for \cite{chang2009rtm} and CRTM. 

We set the hyperparameter $\alpha$ to $5.0$ for all models which leads to consistently good performances for \textit{CRTM} and RTM (identical to the value chosen by~\cite{chang2009rtm} in their evaluations), and set $\rho=2000$ negative examples for the regularization in $\eta$ and $\nu$ estimation.

\subsubsection{Link Prediction}\label{methodo:LP}

We hide 10\% of edges before training models. 

For the four baselines learning documents embeddings, we compare the cosine similarity between hidden pairs and 10 negative examples of unconnected documents. For both RTM and CRTM we use the link function defined at Equation~\ref{eq:rtmexp} and compare the probability of the true link and 10 negative examples. Because CRTM link function doesn't take into account all the content of the source document, Equation~\ref{eq:CRTM} is not suitable for link prediction. Switching from Equation~\ref{eq:CRTM} to \ref{eq:rtmexp} doesn't use the parameter $Q$ learned by CRTM and this modification will be discussed in Section~\ref{LP_ablation}.

We compare CRTM to RTM and 4 recent document network embedding methods specifically designed for link prediction:
\begin{itemize}
    \item \textbf{CRTM}: The context is still defined as the sentence in which the anchor link occurs and we use the same parameters and word-embeddings used for the previous evaluation. We use Equation~\ref{eq:rtmexp} for computing the probability of two documents being linked.
    \item \textbf{RTM}: We use the parameters previously described for anchor prediction.
    \item \textbf{RLE}: The parameter $\lambda$ is set to $0.7$, and RLE outputs document embeddings in dimension $d=160$. The word embeddings are obtained with Skip-Gram with negative sampling, using a window of 15 words and 5 negative examples per true sample, as described in \cite{gourru2020document}.
    \item \textbf{GELD}: We fix $\alpha=0.85$ after gridsearching, and run the model for 40 epochs. GELD outputs vectors in dimension $d=160$. The word embeddings are obtained with Skip-Gram with negative sampling, using a window of 15 words and 5 negative examples per true sample.
    \item \textbf{TADW}: Following~\cite{yang2015tadw}, we reduce the dimension of word vectors to $200$ via SVD decomposition of the TFIDF matrix, select $k = 80$ and $\lambda = 0.2$. We run TADW for 20 epochs.
    \item \textbf{GVNRt}: We run 80 random walks per node of length 40. We fix the window size at 5. GVNRt run for 20 epochs and document embeddings are obtained by concatenating matrix $I$ and $J$.
\end{itemize}

\subsection{Anchor Prediction}
\begin{table*}[ht!]
\caption{Average P@1 and P@5 per corpus for CRTM and baseline (standard deviation in parentheses).\label{tab_patk_rtm_crtmSem}}
\small
\centering
\resizebox{\textwidth}{!}{
    \begin{tabular}{ccccccccccccc}
    \toprule

       \textbf{Corpus}&\multicolumn{2}{c}{\textit{\textbf{Physics}}}&\multicolumn{2}{c}{\textit{\textbf{Society}}}&\multicolumn{2}{c}{\textit{\textbf{Fisica}}}&\multicolumn{2}{c}{\textit{\textbf{Società}}}&\multicolumn{2}{c}{\textit{\textbf{Physik}}}&\multicolumn{2}{c}{\textit{\textbf{Gesellschaft}}}  \\
         \cline{2-13}
         &P@1&P@5&P@1&P@5&P@1&P@5&P@1&P@5&P@1&P@5&P@1&P@5\\
         \midrule
          \textbf{CRTM}& .65 (.02) & \textbf{.86} (.02) & \textbf{.65} (.03)& \textbf{.88} (.01)& .45 (.04)& \textbf{.76} (.03) & .38 (.02)& \textbf{.69} (.01)&.50 (.02)&\textbf{.81} (.001)&.53 (.02)& \textbf{.81} (.01)\\
         \textbf{RTM} & .34 (.03) & .71 (.03) & .32 (.02)& .69 (.02) & .33 (.02) & .72 (.02)& .31 (.02)& .68 (.02) &.39 (.03)& .75 (.02)&.41 (.02)& .75 (.01)\\
         \textbf{TagMe}& \textbf{.71} (.00) & - & .63 (.00)& -& \textbf{.51} (.01)& - & \textbf{.40} (.01)& -&\textbf{.65} (.01) &- &\textbf{.60} (.01)& -\\
         \footnotesize{\textbf{CRTM}}$_1$ & .37 (.02) & .78 (.02)& .36 (.02)& .76 (.01) & .33 (.03)& .73 (.02)& .33 (.02)& .67 (.03)&.39 (.01)& .74 (.01)&.41 (.02)& .76 (.01)\\
         \footnotesize{\textbf{CRTM}}$_U$& .37 (.02)& .76 (.02)&  .36 (.02)& .74 (.01)& .32 (.02)& .70 (.02)& .33 (.02)& \textbf{.69} (.01)&.39 (.02)&.76 (.02)&.42 (.01)&.76 (.01)\\
         \footnotesize{\textbf{CRTM}}$_P$& .37 (.01)& .78 (.01)&  .36 (.01)& .75 (.02)& .34 (.02)& .72 (.01)& .33 (.02)& .67 (.03) &.39 (.03)&.75 (.02)&.41 (.01)& 76 (.01)\\
         \footnotesize{\textbf{CRTM}}$_I$& .37 (.02)& .76 (.02) & .35 (.01) & .74 (.02) & .33 (.03) & .72 (.02) & .33 (.02) & .67 (.03)&.38 (.005)&.74 (.003)&.42 (.01)&.77 (.01)\\  
        \bottomrule
    \end{tabular}}
\end{table*}

Table ~\ref{tab_patk_rtm_crtmSem} reports the average precision at 1 and the precision at 5 for all models over 10 runs, for all five datasets. Figure~\ref{fig_quanti_ROC} shows averaged ROC curves for CRTM and RTM on each datasets.

\subsubsection{Comparison of CRTM with RTM}

\begin{figure}[]
\centering
        \subfigure[Physics]{{
        \begin{tikzpicture}
        \begin{axis}[
        xlabel = {False positive ratio},
        ylabel = {True positive ratio},
        width=0.4\textwidth,
        height=0.3\textwidth,
        xmin=0,
        xmax=1,
        ymin=0,
        ymax=1]
        \addplot[color=blue,mark=, thick] table [y=mean_tpr_AMRTM, x=mean_fpr_AMRTM, col sep=comma] {Physics_ROC.csv};
        \addplot[color=red,mark=, thick, dashed] table [y=mean_tpr_RTM, x=mean_fpr_RTM, col sep=comma] {Physics_ROC.csv};
        \end{axis}
        \end{tikzpicture}}}
     \subfigure[\textit{Society}]{{
        \begin{tikzpicture}
        \begin{axis}[
        xlabel = {False positive ratio},
        width=0.4\textwidth,
        height=0.3\textwidth,
        xmin=0,
        xmax=1,
        ymin=0,
        ymax=1]
        \addplot[color=blue,mark=, thick] table [y=mean_tpr_AMRTM, x=mean_fpr_AMRTM, col sep=comma] {Physics_ROC.csv};
        \addplot[color=red,mark=, thick, dashed] table [y=mean_tpr_RTM, x=mean_fpr_RTM, col sep=comma] {Physics_ROC.csv};
        \end{axis}
        \end{tikzpicture}}}
      
 \caption{Averaged ROC curves of CRTM (blue solid line) and RTM (red dashed line) on the two English corpora.} 
 \label{fig_quanti_ROC}
\end{figure}
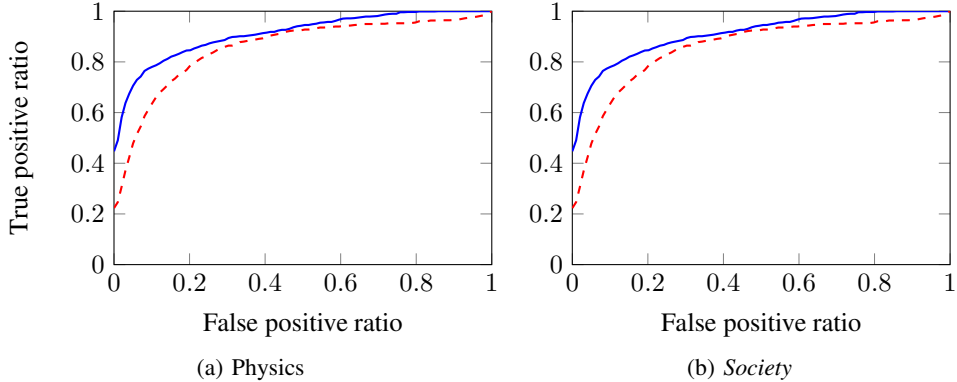

\textit{CRTM} outperforms RTM by a large margin on all datasets. In particular, CRTM exceeds 0.5 in precision at 1 on all English and German datasets, when RTM only score $0.41$ in the most favorable case. This superiority of CRTM is also illustrated by the ROC curves on Figure~\ref{fig_quanti_ROC}.
\textit{CRTM} also outperforms RTM in the precision at 1 on the two Italian corpora, but performs equally as good as RTM in the precision at 5. 
This can be explained in part by the fact that, even though Italian and English documents have the same number of average number words per sentence (about 21.5 in Italian and 19.7 in English), the Italian corpora are smaller, and Italian documents are shorter but with a greater average degree. This may degrade CRTM's ability to generalize over anchor modelization.
A similar explication could be given for the results in the German corpora. German and English datasets tends to have a similar average number of sentences per document, but German sentences are much shorter on our datasets. Furthermore the German language uses declensions and much more surcomposition than English in its vocabulary which could lead to more unique words, and so a weaker topic assignation of those words. 

Overall, CRTM manages to rank the words carrying the hidden links higher than RTM, as evidenced by the ROC curves in Figure~\ref{fig_quanti_ROC}. 

\subsubsection{Comparison of CRTM and TAGME}. TAGME outperforms CRTM in terms of P@1 on the two German corpora, and on the \textit{Physics} one. However CRTM exceeds its competitor on the \textit{Society} by $0.02$ in precision at $1$. For the two Italian datasets TAGME still achieves better performance on P@1, but close to CRTM's one on the \textit{Società} dataset. 
Yet those results may be tempered by the fundamental differences between CTRM and TAGME. As TAGME relies on a predefined dictionary made of previously observed anchors and documents titles, the annotation process is deterministic and TAGME can't infer unseen anchors. On the other hand CRTM aims to infer anchors depending on their topic and context. 
This means that top words predicted by CTRM are not necessarily inappropriate, even though they didn't match the exact anchor of the removed hyperlink.
By saying this we could mitigate those results and advance that CRTM's P@5 performance, exceeding by a large margin TAGME's P@1 on all datasets, can lead to a better diversity of links. We further illustrate this point in Section \ref{usecase}. 

\begin{table}[ht]
\caption{Average wall-clock runtime in seconds (time to train the embeddings in parentheses when applicable).\label{tab_time}}
\centering
    \begin{tabular}{lcc}
    \toprule
    &\textbf{Average runtime (s)}&\textbf{Average runtime/iter (s)}\\
    \midrule
    \textbf{CRTM}&$2.2 \times 10^{2} (+10^{2})$&28.4\\
    \textbf{RTM}&$4.6\times 10^{2}$& 11.4\\
    \textbf{CRTM}$_1$&$14.3\times 10^{2}$&12.2\\
    \textbf{CRTM}$_U$&$6.1\times 10^{2}$&18.8\\
    \textbf{CRTM}$_P$&$6.4\times 10^{2}$&18.2\\
    \textbf{CRTM}$_I$&$5.1\times 10^2 (+10^{2})$&12.73\\
    \bottomrule
    \end{tabular}{}
\end{table}

\subsubsection{Execution time}

We report in Table \ref{tab_time} the average wall-clock execution time for all models, across all datasets. All models are coded in Python 3 and trained using a single core on a computer with an \textit{Intel i7-6700K} CPU and 64GiB of RAM. We notice that all the average runtimes are in $O(10^2)$ seconds, including the time required to train the word embeddings in \textit{CRTM}. Interestingly, while an iteration takes longer to compute in \textit{CRTM}, more than twice the time needed for RTM, CRTM needs fewer iterations to converge and thus ends up running faster than RTM.

\subsubsection{Ablation Study}\label{seq:ablation_AC}

In this section, we study the performance of CRTM and its variants to show the importance of each of its components.

\paragraph{Impact of the attention mechanism.} To judge the impact that the attention mechanism has on the performance of CRTM, we must compare it with these of CRTM$_1$ (no context), CRTM$_U$ (simple average) and CRTM$_P$ (Gaussian smoothing around the word's position). CRTM$_1$ has slightly better performance than RTM and is largely outperformed by CRTM. This tends to show that restricting the context to the sole word carrying the link is often too drastic, which can also explain why RTM struggle to match the performance of CRTM. CRTM$_U$ manages to equal CRTM on the \textit{Società} dataset in precision at 5. However it is consistently outperformed by CRTM in all the other settings, and in particular in precision at 1. This suggests that simply averaging the per-word topic assignments is sub-optimal. On the other hand, CRTM$_P$ manages to improve over RTM, with a relative gain of up to 7\% in the most favorable cases. This seems to indicate that the topic assignments of the words closest to the one carrying the hyperlink are more important. Yet, CRTM$_P$ is still outperformed by CRTM in the precision at 1 and 5 on all datasets, which shows that the attention mechanism is an important component of CRTM.

\paragraph{Impact of $Q$.}\label{AP_Q} Here we compare the performance of CRTM with the performance of CRTM$_I$ (where $Q = I_K$). CRTM consistently outperformed CRTM$_I$ in precision at 1 and 5 on the six considered corpora. We can interpret that by saying that learning new representations is beneficial.

\subsection{Link Prediction}
\begin{table*}[ht!]
    \caption{AUC for link prediction (standard deviation in parentheses)}
    \label{tab:LP_AUC}
    \centering
    \begin{tabular}{lllllll}
    \toprule
                & \textit{\textbf{Physics}} & \textit{\textbf{Society}} & \textit{\textbf{Fisica}} & \textit{\textbf{Società}} & \textit{\textbf{Physik}}&\textit{\textbf{Gesellschaft}}  \\
    \midrule
       \textbf{CRTM } & .87 (.004)&  .87 (.003)    & .81 (.01)    & .78 (.02)& .82 (.002)   &  .80 (.01) \\ 
       \textbf{RTM }& .85 (.02)&  .86 (.01)   & .80 (.01)    & .78 (.03)    & .82 (.005)  &  .78 (.01)  \\
       \textbf{RLE  }   &  .89 (.002)&  .91 (.003)   & .90 (.002)    & .89 (.001) & .86 (.002)  & .88 (.001)   \\
       \textbf{GELD } & .89 (.001)&  .91 (.001)   & .89 (.001)    & .88 (.002) & .87 (.001) &  .85 (.001)  \\
       \textbf{TADW }& .83 (.003)&  .82 (.003)  & .80 (.005)    & .73 (.001) &.67 (.003) & .65 (.004)   \\
       \textbf{GVNR-t}  & .97 (.03)&   .96 (.04)   & .95 (.001)     & .95 (.04)     & .96 (.03)& .96 (.04) \\
       \bottomrule
    \end{tabular}

\end{table*}

Table~\ref{tab:LP_AUC} reports the AUC scores for all datasets. We run each model five times, on each dataset, and report mean AUC with standard deviation.

\paragraph{Results analysis} CRTM beats TADW by a clear margin on five of six datasets. CRTM's performance matches RTM or slightly improves upon it, while coming close to RLE and GELD on English datasets. GVNR-t clearly performs better than all methods. Still, we observe that CRTM achieves consistent performances, with at least 0.80 AUC in all cases.
It is worth noting that GVNR-t has a training time of several hours on our datasets (due to the computational costs of random walks), while CRTM is trained in less than 20 minutes on the same computer. 
This experiment shows that taking into account the context of anchor links at least doesn't degrade CRTM results in link prediction w.r.t RTM, and leads in some cases to a minor performance boost.

\label{LP_ablation}
\paragraph{Impact of Q}
As mention in section \ref{methodo:LP}, CRTM's link function is not suitable for link prediction as it requires knowing the anchor. Therefore once we've trained CRTM we substitute its link function with the simpler RTM's link function to do link prediction, thus omitting the transformation $Q$.
We further investigate the impact of this parameter by introducing the transformation $Q$ in the link function of RTM.

The overall average AUC in link prediction on the six datasets falls to $0.63$, while it is about $0.80$ without using the parameter $Q$. This result, coupled with Section~\ref{AP_Q}'s conclusion, suggest that (i) CRTM learns topics that are useful for link prediction and that (ii) $Q$ confers it the ability to efficiently recombine topics so that they are suited to anchor prediction. This means that a single CRTM model could solve both the link prediction (even though it doesn't match the performance of the most up-to-date techniques) and anchor prediction tasks, by simply changing the link function at prediction time.

\subsection{Case Study}\label{usecase}

Here we show the anchors that CRTM is able to automatically detect, given a pair of source and target documents. To this end, CRTM was trained with all hyperlinks removed in the source documents, to prevent it from simply listing anchors seen during training. These anchors could be used by writers, to automatically insert hyperlinks towards related pages. They could also be useful for readers, as they could serve as contextual hyperlinks, linking the page they are reading with those they've previously read. We also report the anchors found by RTM, to highlight how those found by CRTM are more relevant.

As an example, Table~\ref{qualitative_exp_semiconductor} shows the five most likely anchors, given the page about ``Semiconductor'' as the source, and the page about ``Transistor'' as the target, and Table~\ref{qualitative_exp_computer} shows the five most likely anchors, given the page about ``Computer'' as the source, and the page about ``Transistor'' as the target. We note that CRTM always rank the word transistor first, the most natural word to be an anchor. RTM actually manages to identify it as an important word too, but ranks it lower. CRTM also ranks MOSFET at the second or third place, which is a type of transistor. In addition, we observe that RTM predicts some less interesting words, such as \textit{loom}, \textit{fabricated} or \textit{enable}, while CRTM highlights relevant but less obvious terms, like \textit{Moore} (from the Moore's law).
\begin{table}[!h]
\caption{Five most likely anchors in the ``Semiconductor'' page, connected to the ``Transistor'' page.}
\label{qualitative_exp_semiconductor}
\centering
\begin{tabular}{p{.9\linewidth}}
\hline
\textbf{Transistor}, A transistor is a semiconductor device used to amplify or switch electronic signals and electrical power. [\dots] \\
\hline
\end{tabular}
\begin{tabular}{|c||c|}
\multicolumn{2}{c}{\textuparrow}\\
\multicolumn{2}{c}{\textbf{Semiconductor}}\\
\hline
\textbf{CRTM}& \textbf{RTM}  \\
\hline
transistor&loom\\
MOSFET&MOSFET\\
semiconductor&transistor\\
circuit&enable\\
Moore&circuit\\
\hline
\end{tabular}
\end{table}

\begin{table}[!ht]
\caption{Five most likely anchors in the ``Computer'' page, connected to the ``Transistor'' page.}
\label{qualitative_exp_computer}
\centering
\begin{tabular}{p{.9\linewidth}}
\hline
\textbf{Transistor}, A transistor is a semiconductor device used to amplify or switch electronic signals and electrical power. [\dots] \\
\hline
\end{tabular}
\begin{tabular}{|c||c|}
\multicolumn{2}{c}{\textuparrow}\\
\multicolumn{2}{c}{\textbf{Computer}}\\
\hline
\textbf{CRTM}& \textbf{RTM}  \\
\hline
transistor&staircase\\
electron&atom\\
MOSFET&dopant\\
dopant&fabricated\\
electrical&ability\\
\hline
\end{tabular}
\end{table}

\section{Conclusion and future work}\label{CCL}
We have presented the Contextualized Relational Topic Model, CRTM, a probabilistic modeling framework to infer latent topics in networks of documents, that explicitly accounts for the locations of the links in the text. We've experimentally shown the relevancy of our approach through a quantitative evaluation based on several Wikipedia datasets, in English, Italian and German. We've also shown that CRTM has a competitive runtime, which makes it usable in practice to solve tasks akin to anchor prediction without relying on external information like a knowledge graph.
We also demonstrated that taking anchor links into account doesn't degrade the model's performance in link prediction.
From a qualitative point of view, we've illustrated how CRTM can assist knowledge bases contributors while they specify anchor links after writing. 
In future work, we'd like to investigate more complex link functions based upon a more sophisticated modeling of topics and documents, and extend our work to more recent works, like Graph Neural Networks and Neural Topic Models.

\bibliographystyle{unsrtnat}
\bibliography{references}  






\end{document}